\documentclass[11pt]{article}
\usepackage{acl}
\usepackage{times}
\usepackage{latexsym}
\usepackage{booktabs} 
\usepackage{makecell} 
\usepackage{caption} 
\usepackage{multirow} 
\usepackage{xcolor}

\usepackage[T1]{fontenc}

\usepackage[utf8]{inputenc}
\usepackage{graphicx}
\usepackage{booktabs}
\usepackage{makecell}
\usepackage{multirow}
\usepackage{multicol}
\usepackage{enumitem}
\usepackage{float}
\usepackage{xcolor}
\usepackage{blindtext}
\usepackage{tcolorbox}
\usepackage{tabulary}

\definecolor{blizzardblue}{rgb}{0.97, 0.97, 1.0}

\usepackage{microtype}

\title{FlowVQA: Mapping Multimodal Logic in \\Visual Question Answering with Flowcharts}

\author{
  Shubhankar Singh\textsuperscript{1}\textsuperscript{†}, 
  Purvi Chaurasia\textsuperscript{2}\textsuperscript{†},
  \textbf{Yerram Varun}\textsuperscript{3}\thanks{~,~† ~contributed equally, \ddag~primary mentor \& corresponding author} , \textbf{Pranshu Pandya}\textsuperscript{4}\textsuperscript{*},\\
\textbf{Vatsal Gupta}\textsuperscript{4}\textsuperscript{*},
   \textbf{Vivek Gupta}\textsuperscript{5}\textsuperscript{\ddag}, \textbf{Dan Roth}\textsuperscript{5} \\
  \textsuperscript{1}Mercer Mettl,
  \textsuperscript{2}IGDTUW New Delhi,
  \textsuperscript{3}Google Research\\
  \textsuperscript{4}Indian Institute of Technology Guwahati, 
  \textsuperscript{5}University of Pennsylvania \\
    \texttt{\small shubhankar.singh@mercer.com}, \texttt{\small purvi069btcsai21@igdtuw.ac.in},  \\
  \texttt{\small vyerram@google.com}, \texttt{\small \{p.pandya,g.vatsal\}@iitg.ac.in},  \\
  \texttt{\small \{gvivek, danroth\}@seas.upenn.edu} \\
}

\newcommand{\paraendspace}{\vspace{1.0mm}}

\begin{document}
\maketitle
\begin{abstract}
Existing benchmarks for visual question answering lack in visual grounding and complexity, particularly in evaluating spatial reasoning skills. We introduce FlowVQA, a novel benchmark aimed at assessing the capabilities of visual question-answering multimodal language models in reasoning with flowcharts as visual contexts. FlowVQA comprises 2,272 carefully generated and human-verified flowchart images from three distinct content sources, along with 22,413 diverse question-answer pairs, to test a spectrum of reasoning tasks, including information localization, decision-making, and logical progression. We conduct a thorough baseline evaluation on a suite of both open-source and proprietary multimodal language models using various strategies, followed by an analysis of directional bias. The results underscore the benchmark's potential as a vital tool for advancing the field of multimodal modeling, providing a focused and challenging environment for enhancing model performance in visual and logical reasoning tasks.

\end{abstract}

\section{Introduction}
\label{sec:intro}
\begin{figure}[h]
\vspace{2mm}
\centering 
\includegraphics[width=0.85\columnwidth]{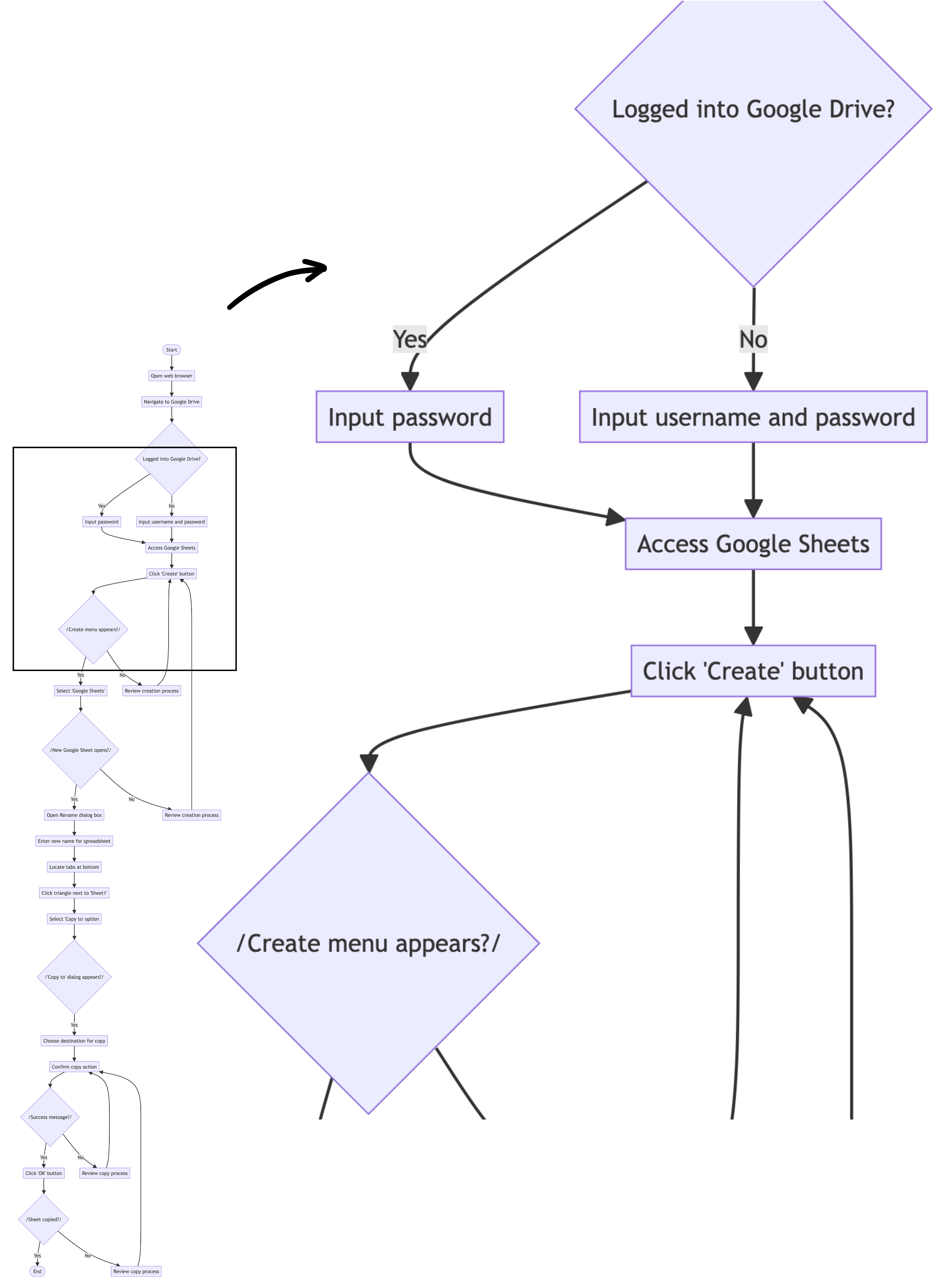} 
  \small{
\begin{table}[H]
  \centering
  \small
  \vspace{-0.5em}
  \begin{tabular}{p{0.95\columnwidth}} 
    Q. Derek wants to ensure that the sheet was successfully copied before reporting back to Melissa. What should Derek see or do next to ensure the task was completed correctly?
    
    \vspace{0.5em}
    A. He should look for a success message and dismiss the dialogue by clicking 'OK'.
  \end{tabular}
  \label{tab:qa}
\end{table}
  }
  \vspace{-1.5em}
\caption{\small A zoomed-in section of a flowchart in our resource set along with a correspnding QA pair. wiki00203: "How To Convert an Old Google Spreadsheet to Google Sheets." A detailed example of a flowchart along with its question-answer pairs is outlined in Appendix~\ref{apex:flowchart-qa-example}. } 
\label{fig:mainfig} 
  \vspace{-2.0em}
\end{figure}

Since the inception of Vision Language Models (VLMs), tasks and benchmarks for visual question answering (VQA) and reasoning have received significant attention. Most benchmarks evaluate pre-trained model extraction capabilities, neglecting their ability to comprehend complex spatial relationships and visual logical reasoning. Research on spacial path following or visual sequential reasoning in VLMs is limited. Current benchmarks for assessing VLM reasoning abilities mainly fall under VQA, a task formalized by \citet{goyal2017making}, which involves generating responses to questions based on a given image. These works evaluate the spatial reasoning and visual information extraction abilities of VLMs.

Visual Grounding (VG) of a visual question-answering system evaluates models' abilities to attribute their generations to different image regions referenced in the query \cite{reich-etal-2023-measuring}. The absence of VG has been a frequent issue among the current VQA systems, manifesting in an over-reliance on irrelevant parts of images or a complete disregard for the visual modality. Existing benchmarks~\cite{yue2023mmmu} require models to rely on pre-trained knowledge to answer queries posed on image contexts. In this work, we aim to test the capabilities of VLMs in following visual information without any pre-existing knowledge. 
To accomplish this, we delve into the realm of flowcharts, as depicted in figure~\ref{fig:mainfig}, which entail intricate structural configurations and path reconstruction, a considerably more challenging task compared to mere image comprehension.

Flowcharts \textbf{emphasize sequential and logical reasoning}, as they necessitate traversal of steps or decisions in a specific sequence. Flowcharts are \textbf{inherently visual}, and provide a clear and structured method for representing processes, decision paths, and flows. Unlike traditional text, which flows linearly, flowcharts require an understanding of \textbf{directional logic}; their flow is often multi-directional, representing various paths that can be taken based on certain conditions or decisions. Despite being long and complex, flowcharts have \emph{compact}, \emph{systematic} representations and provide insights regarding information in a step-by-step manner. 
\paraendspace

In this paper, we set out to answer a crucial question: \emph{"Can modern Vision Language Models effectively handle challenges that demand understanding both structural and semantic aspects, along with capturing macroscopic and granular context within visually complex yet straightforward flowcharts?"} To tackle this question, we introduce \textbf{FlowVQA}, a novel benchmark comprising intricate structural and path-based questions posed on lengthy flowchart images. We propose a novel approach to Visual Question Answering (VQA) on Flowchart tailored for VLM, with a focus on harnessing flowcharts as the primary contextual framework for visual logic and spatial reasoning.
\paraendspace

 \textbf{FlowVQA} consists of 2,272 Mermaid.js flowchart scripts generated with human input, sourced from process workflow articles like Instructables and WikiHow, as well as Code. Accompanying these are 22,413 Q/A pairs covering various reasoning skills like information localization, fact retrieval, scenario deductions, flow reasoning, and topological understanding. The creation process involves a meticulous multi-step machine generation and human verification to discard up to 51\% of samples, ensuring they meet high standards of challenge, coherence, and insightfulness. This rigorous process grounds the flowchart reasoning in textual domain, enriching the visual task complexity. Extensive experimentation revealed that both closed and open-source Vision Language Models (VLMs), equipped with a range of prompting strategies and fine-tuning techniques, struggled to execute visual and spatial reasoning tasks within the FlowVQA dataset. Moreover, our findings highlighted a directional bias and non-uniform performance pattern across flowcharts of varying lengths exhibited by these VLMs. Our contributions are the following:
\begin{itemize} [leftmargin=*, noitemsep]
    \setlength{\itemsep}{0.75pt} 
    \item Introduction of VQA for FlowCharts, focusing on visual logic and spatial reasoning, filling a gap in previous benchmarks.
    
    \item Development of a detailed framework for generating intricate VQA samples transitioning from text to visual domains, ensuring quality, complexity, and accuracy via rigorous verification.
    
    \item Introducing the novel benchmark FlowVQA, consisting of 2,272 high-quality Flowchart Images and 22,413 Q/A samples spanning four distinct question types, created using the framework.
    
    \item Thorough evaluation of closed and open-source VLMs, employing various prompting strategies and fine-tuning methods. An analysis of directional bias and non-uniform performance across different flowchart lengths.

\end{itemize}

The FlowVQA dataset, along with modeling and evaluation scripts, generation pipeline and prompts, and the human verification platform, can be accessed at \href{https://flowvqa.github.io/}{https://flowvqa.github.io/}.

\section{Proposed FlowVQA Resource}
\label{sec:our-flowvqa-resource}

In this section, we will see the details of the construction of \textbf{FlowVQA}. We outline the process of collection of raw data from wild sources, multi-step generation of mermaid scripts and flowchart images and complex Q/A creation.

\begin{figure*}[!ht]
    \centering
    \includegraphics[width=0.85\textwidth]{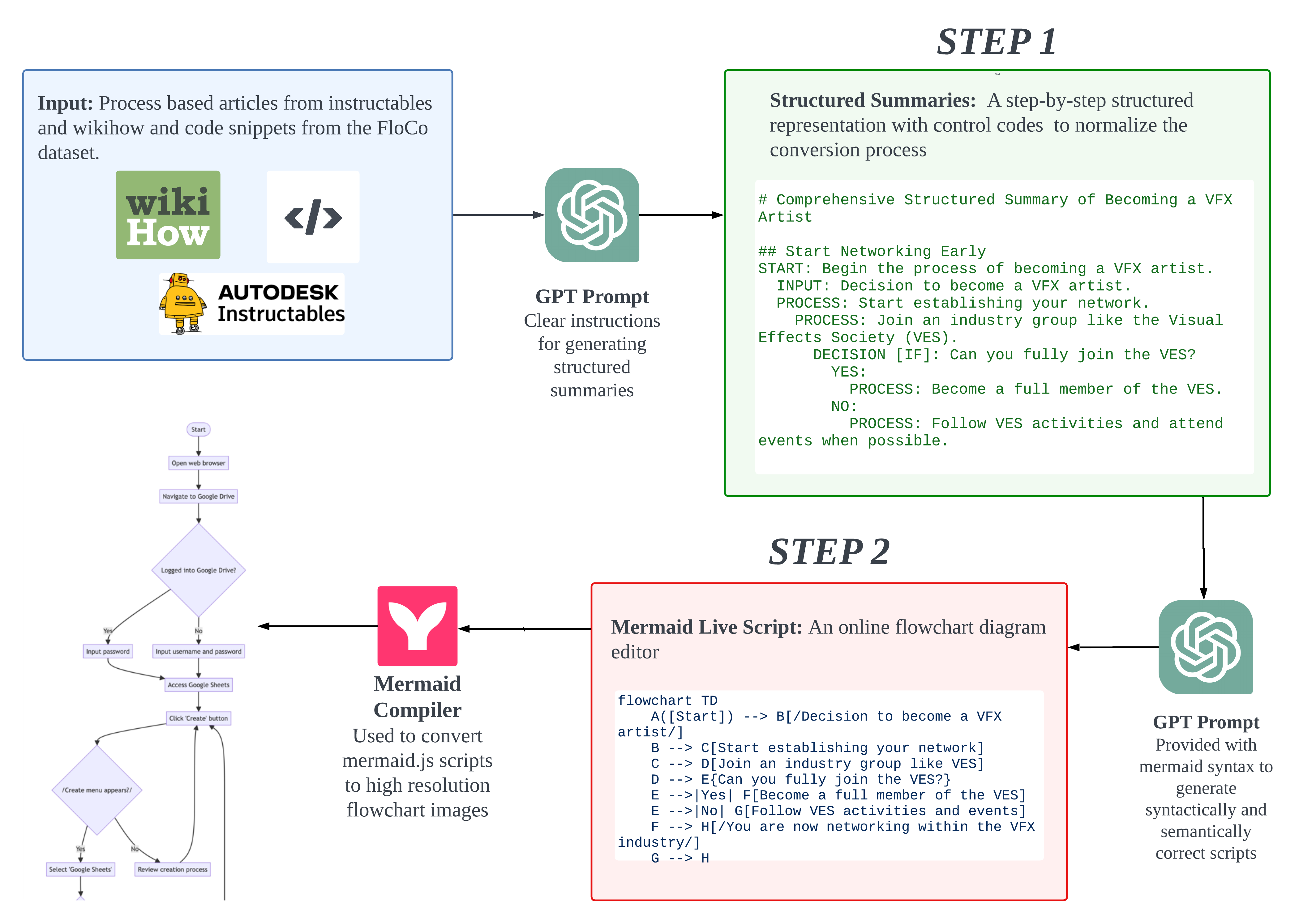}
    \caption{\small Our dataset's generation pipeline encompasses the creation of flowcharts. As previously outlined, we employ a comprehensive two-step process to derive high-quality flowcharts from source texts. Additionally, to guarantee accurate generation, a cross-verification mechanism is implemented.}
    \label{fig:flowchart-generation-pipeline}
\end{figure*} 

\subsection{Flowchart Sources}
We draw input texts from three primary sources: \textbf{WikiHow} articles, \textbf{Instructables} DIY blogs, and \textbf{FloCo}~\cite{DBLP:conf/icdar/ShuklaGKYM23} code snippets. WikiHow and Instructables provide \textit{step-by-step instructions} for everyday tasks, while the FloCo dataset, \textit{a flowchart-to-code} resource, features low-complexity code samples. We categorize all the WikiHow articles, Instructables DIY based on the domains of these articles. FloCo code snippets are categorized into \emph{code} category. The distribution across categories is outlined in Appendix~\ref{apex:dataset-distribution}.

We manually select high-quality code snippets from FloCo to ensure uniformity in our pipeline across all text sources. FloCo image samples enable us to iteratively compare the generated flowcharts with the original samples. This step was crucial as it helped perfect our prompts and allow applicability to the WikiHow and Instructables set. We sample 1,268 WikiHow articles, 789 Instructables blogs, and 475 FloCo examples as an input to our human verification pipeline.

\begin{table}[!ht]
\centering
\small
\addtolength{\tabcolsep}{-0.2em}
\begin{tabular}{lccc}
\toprule
 & \textbf{WikiHow} & \textbf{Instructables} & \textbf{FloCo} \\
\midrule
Source Texts & 1,914 & 943 & 700 \\
Mermaid.js Scripts & 1,500 & 792 & 575 \\
\bottomrule
\end{tabular}
\vspace{-0.5em}
\caption{\small FlowVQA Generation resources.}
\vspace{-1.0em}
\label{table:flowchart-generation-errors}
\end{table}

\textbf{Generation and Filteration}. GPT-4 based data generation of data and benchmarks is prevalent \cite{DBLP:journals/corr/abs-2311-16483} in prior works. Machine generation method for flowcharts and Q/A has several advantages to crowdsourcing: (i) The complex and intricate process of creating flowcharts and Q/A pairs constitutes a laborious, efficient and a time-intensive task for human workers, (ii) Using GPT-4 for the generation of structured representations and subsequent conversion into flowcharts and Q/A pairs enables rapid scaling, (iii) The Stochastic nature of LLMs helps in the creation of an unbiased and diverse Q/A dataset. To produce Flowchart and Q/A Samples, we employ an automated 'generate-and-test' approach, where we exhaustively generate questions of multiple reasoning types and apply rigorous filtration to maintain the quality, hardness, and correctness of samples through effective prompting with GPT-4. Our meticulous verification through experts and rubrics, along with our custom-built annotation platform, ensures a thorough and impartial evaluation of both flowcharts and Q/A pairs.

\begin{table*}[!h]
\centering

\small
\begin{tabular}{lccccccccc}

\toprule
\textbf{Source} & \textbf{\# Samples} & \textbf{Avg. NPF} & \textbf{Avg. EPF} & \textbf{Avg. Width} & \textbf{Avg. Height} & \textbf{Ratio} & \# \textbf{Qs.}\\
\midrule
Wikihow\textbf{ }& 1,121 & 21.83 & 24.04 & 1568.0 & 5551.81 & 1 : 3.54 & 11,957  \\
Instructables & 701 & 19.76 & 21.18 & 1568.0 & 6629.80 & 1 : 4.23 & 6,893    \\
Code & 450 & 9.87 & 10.85 & 1568.0 & 2738.15 & 1 : 1.75 & 3,563   \\
\midrule
Full & 2,272 & 18.82 & 20.54 & 1568.0 & 5327.13 & 1 : 3.40 & 22,413   \\

\bottomrule
\end{tabular}
\vspace{-0.5em}
\caption{\small FlowVQA Source-wise Statistics: Number of Flowchart Samples, Average Nodes Per Flowchart, Average Edges per Flowchart, Average Image Width (Pixels), Average Image Height (Pixels), Aspect Ratio and Number of Questions. (The flowchart image render is set for a constant width factor)}
\vspace{-1.0em}
\label{table:source-wise-stats}
\end{table*}

\subsection{Flowchart Generation} Our primary supposition for flowchart creation is that \textit{any process-based workflow, regardless of domain, can be converted to a flowchart which highlights key aspects of the process in a detailed step-by-step fashion.} We treat the conversion of source article to flowchart Mermaid Scripts as a two-step soft-syntax summarization task. Ideally, we would use real-world flowcharts from external sources such as books and documents, but the availability of such structured data is extremely sparse. Initially, we aimed to use real-life flowcharts, but the scarcity of standardized flowcharts and the lack of sufficient open-source examples made it unfeasible to create a dataset as comprehensive as ours. We decouple the structured summarization into a flowchart script to implement this two-step process. 

\textit{\textbf{First Step}}. We query GPT-4 with the source text to generate a step-by-step structured representation of the text annotated with functional control tags (e.g., “START,” “PROCESS,” “DECISION”). This step converts the source text into a tagged textual representation suitable for converting into mermaid flowchart scripts. For FloCo-sourced texts, we generate pseudocode for the code scripts as the input to the next step. 

\textit{\textbf{Second Step}}. In this step, we generate the Mermaid.js flowchart script(top-down) using the output of the \emph{first step} by querying GPT-4 with a template Mermaid.js script. The control tags facilitate mapping the steps to the node types used in the script. Constraining points are provided alongside both prompts for improved normalization. The Mermaid.js scripts are then compiled to create high-resolution PNG images. 

Table~\ref{table:flowchart-generation-errors} represents the number of samples after the two-step conversion process. We exclude the scripts and representations with minor syntactical and rendering errors. Figure \ref{fig:flowchart-generation-pipeline} showcases the generation pipeline of the flowcharts in our dataset. Appendix~\ref{apex:flowprompts-for-generation}
lists the prompts used in \textit{first step} and \textit{second step}.

\subsection{Question Answer Creation}
\label{subsec:qa-creation}
We curate four question types designed to analyze and test different aspects: Fact Retrieval, Applied Scenario, Flow Referential and Topological Question and Answer. First three can be broadly categorized under granular flowchart comprehension while topological tests structural information. 

\textit{\textbf{T1. Fact Retrieval:}} These simple questions involve the localization and retrieval of direct factual information from flowchart's nodes. Despite being simple, they still necessitate image analysis and retrieving relevant cues that localize the final answer.

\textit{\textbf{T2. Applied Scenario:}} These questions describe a real-life scenario and test the models' application of the flowchart to a practical problem. These questions capture reasoning skills used by humans parsing flowcharts in day-to-day life. It leads to interesting puzzle-like word problems that test the understanding of decision steps, content, and reasoning in the presence of distractor context, which needs to be filtered to understand the question.

\textit{\textbf{T3. Flow Referential:}} In these questions, A random sub-graph/section of the flowchart, usually involving a decision node, is considered, and a question is formulated on backward-forward flow with decision-based logic. It assesses granular path dynamics in a flowchart.

\textit{\textbf{T4. Topological:}} This question type addresses the larger topology of a flowchart, requiring analysis of the flowchart at a more macroscopic level to give an answer related to the structural topology of the graph. These questions are created by parsing Mermaid.js scripts to convert them into an adjacency matrix representing the flowchart in the form of a graph. It generates template-based questions that usually have quantitative correct answers.

\begin{figure*}
    \centering
    \includegraphics[width=0.95\textwidth]{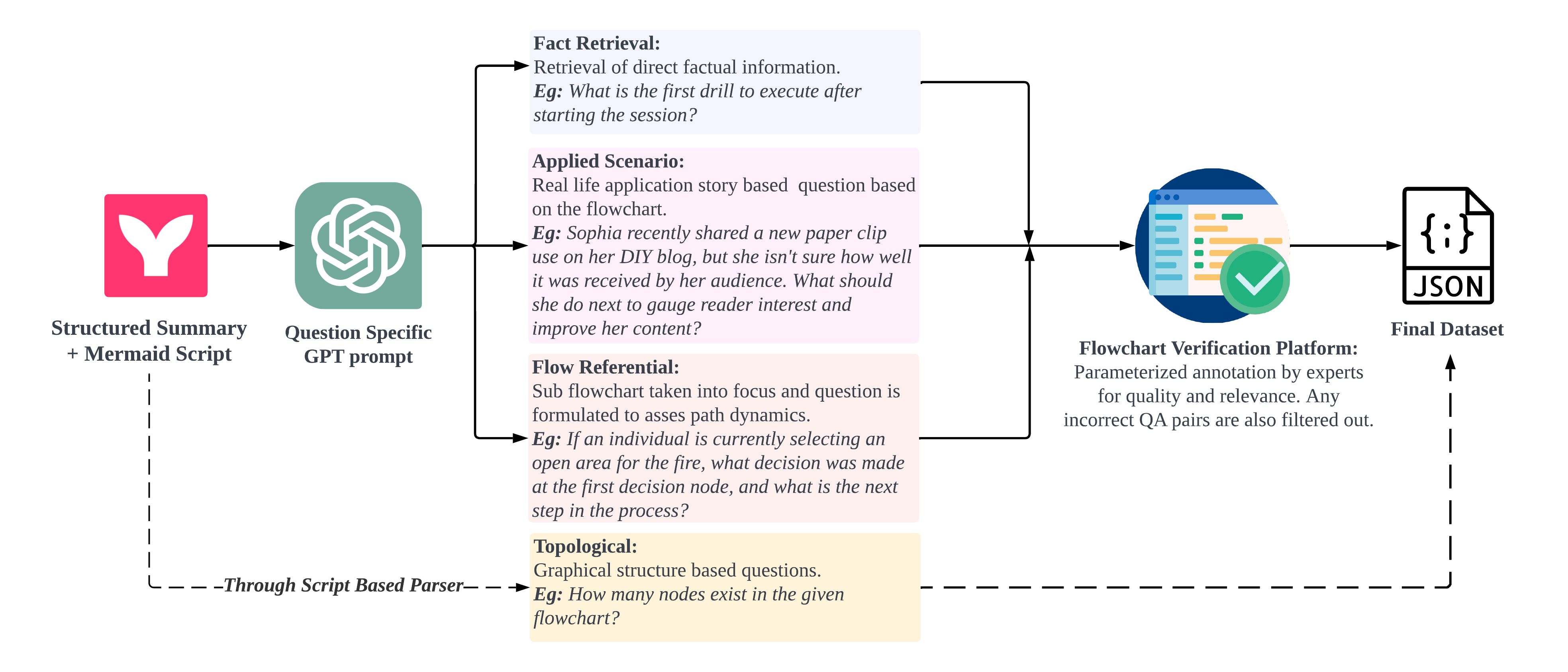}
    \caption{\small Our dataset incorporates a question creation pipeline tailored to accommodate various question types. As previously noted, each question type undergoes generation via a carefully crafted prompt, meticulously designed to achieve the specific objectives associated with that type of question}
    \label{fig:qa-generation-pipeline} 
\end{figure*}

\begin{table}[!htbp]
\vspace{-0.5em}
\small
\centering 
\begin{tabular}{@{}lcccc@{}} 
\toprule
 \textbf{Statistics}& & \multicolumn{1}{c}{\textbf{Train}} & \multicolumn{1}{c}{\textbf{Test}} & \multicolumn{1}{c}{\textbf{Total}} \\
\midrule
\multicolumn{2}{l}{\textbf{Total Flowcharts}} & 1,319 & 953 & 2,272 \\
\multicolumn{2}{l}{\textbf{Avg. Nodes}} & 18.63 & 19.09 & 18.82 \\
\midrule
\multirow{4}{*}{\textbf{QA}} & \multicolumn{1}{l}{Fact Retrieval} & 2,654 & 1,878 & 4,532 \\
& \multicolumn{1}{l}{Applied Scenario} & 2,640 & 1,936 & 4,576 \\
& \multicolumn{1}{l}{Flow Referential} & 2,128 & 1,585 &  3,713 \\
& \multicolumn{1}{l}{Topological} & 5,516 & 4,076 & 9,592 \\
\midrule
\multicolumn{2}{l}{\textbf{Total QA}} & 12,938 & 9,475 & 22,413 \\
\bottomrule
\end{tabular}
\vspace{-0.5em}
\caption{\small QA Resource Split Statistics.} 
\label{table:dataset-split-stats}
\vspace{-1.0em}
\end{table}

\textbf{Q/A Generation}. We construct a prompt to query GPT-4 using the tagged textual representation, Mermaid.js script and text-only few-shot examples to generate high quality Q/A pairs of types, T1, T2 and T3 (listed in Appendix~\ref{apex:qgprompts-for-generation}). For each question, we generate three paraphrased gold answers, which allows us to evaluate models irrespective of their generation syntactics and semantics. As part of text-only few-shot examples we pass a variety of creative high-quality examples. Topological Q/A pairs (T4) are generated by parsing the Mermaid script, converting the graph into an adjacency matrix, and creating template-based questions. Answers are usually quantitative. After formulating the template-based answers, we obtain two additional paraphrased answers for each template answer to achieve three gold-standard answers, thus maintaining the standard with the other question type for three gold short answers.

\subsection{Human Verification Pipeline and Platform}
To ensure strong validity of our work, we establish a robust human verification pipeline for our models and flowcharts. All generated outputs for flowcharts and subsequent Q/A pairs undergo a rigorous quality check by a team of five expert annotators. As we adhere to a "Generate-and-test" paradigm (section~\ref{sec:our-flowvqa-resource}), we provide detailed rubrics for both flowchart and Q/A pair verification and annotation, with parameters such as logical flow, complexity, context alignment and more, for flowcharts and Q/A pairs which allow the annotators be strict and thorough. To assist with their work and eliminate any bias and stress, we also provide them with a detailed, custom-built annotation platform to provide scores, filter out, etc. This custom platform enables parallel viewing. 
\paraendspace

\begin{table}[!htbp]
\vspace{-0.5em}
\centering
\small
\begin{tabular}{lccccc}
\toprule
 & \textbf{\# Samples} & \textbf{\# T1} & \textbf{\# T2} & \textbf{\# T3} \\
\midrule
Pre & 2,532 & 8,932 & 9,138 & 7,262  \\
Post & 2,272 & 4,532 & 4,576 & 3,713  \\
\midrule
\% drop & 10.3\% & 49.3\% & 50\% & 48.9\% \\
\bottomrule
\end{tabular}
\vspace{-0.5em}
\caption{\small FlowVQA Annotation-based filtering stats pre and post-verification and filtration for number of flowchart samples and QA Types \textit{T1}, \textit{T2} and \textit{T3}}.
\label{table:flowchart-annotations-filtered-stats}
\vspace{-1.0em}
\end{table}

\textbf{Annotation Platform}. Our custom-built annotation platform consists of UI, where we pass the flowchart and Q/A pairs together so they can be viewed simultaneously. The annotators provide quality scores\footnote{captures the consistency, correctness and complexity.} for all components of the dataset and a final holistic score\footnote{captures the relevancy between the components.}. We filter out flowcharts below a fixed quality threshold and Q/A pairs which rate below average. Topological questions are not passed into the platform as they are hard-template-based and obtained via scripting. All verification products are verified with two separate supervising experts who ensure the quality of annotations is consistent and scores remain unbiased. The verification lasts for ten days from start to end. 

The final samples, see Appendix~\ref{apex:flowchart-qa-example}, ensure appropriate complexity and correctness of flowcharts, questions and corresponding answers. Figure \ref{fig:qa-generation-pipeline} showcases the complete question-answer generation pipeline used to create the dataset.

\section{Experimental Evaluation}
\label{sec:experiment_evaluation}
We address the following research questions through our experiments: 

\vspace{0.5em}
\textbf{\textit{RQ1.}} Does the introduced visual multimodal dataset present a significant challenge to current multimodal language learning models (VLMs), and can it provide valuable insights that could contribute to their future advancement? 

\vspace{0.5em}
\textbf{\textit{RQ2.}} Is the efficacy of VLMs influenced by factors such as (a) the source of flowcharts, (b) the type of questions posed, and (c) the level of complexity inherent in the flowcharts? 

\vspace{0.5em}
\textbf{\textit{RQ3.}} Are there ways to enhance the performance of visual question answering tasks related to flowcharts through the use of specific directives tailored to flowcharts? Moreover, does the process of fine-tuning these models with the train split of FlowVQA dataset improve their proficiency in handling questions tied to flowchart-based data?

\vspace{0.5em}
\textbf{\textit{RQ4.}} Is there an observable directional bias in existing VLMs when they are applied to flowchart analysis?

\vspace{0.5em}
{\bf Limitations of Smaller Models}. FlowVQA represents a complex multimodal challenge that requires visual logic and reasoning across large-scale high-resolution images. In our assessment of several widely utilized open-source multimodal language learning models (VLMs) -- including \textbf{LLaVA}~\cite{DBLP:journals/corr/abs-2304-08485}, \textbf{OpenFlamingo}~\cite{DBLP:journals/corr/abs-2308-01390}, \textbf{BLiPv2}~\cite{DBLP:conf/icml/0008LSH23}, \textbf{mPLUG-OWL}~\cite{DBLP:journals/corr/abs-2304-14178}, Sphinx~\cite{DBLP:journals/corr/abs-2311-07575} — we observe that their performance on our test dataset \textbf{is notably subpar (<10\%)}. These multimodal language learning models (VLMs) lack a sizable vision encoder, leading to the internal distortion of flowchart images with high aspect ratios when passed into the vision encoder. Furthermore, even if they can interpret the image a bit, their inadequate reasoning abilities render them extremely ineffective for any further analysis utilizing this resource.

\vspace{0.5em}
{\bf Models for Comparison}. We perform evaluations on FlowVQA with \textbf{five} different VLMs. We employ \textbf{GPT-4V}~\cite{DBLP:journals/corr/abs-2303-08774} and \textbf{Gemini Pro}~\cite{DBLP:journals/corr/abs-2312-11805}\footnote{We use the preview version for Gemini Pro at Vertex API \citep{geminicite}. Gemini Ultra is/was not made public yet.} to test the visual understanding capabilities of best proprietary (closed) models available. We also employ three open-source models.
\textbf{CogAgent-VQA}~\cite{DBLP:journals/corr/abs-2312-08914} is an 18-
billion-parameter visual language model (VLM) specializing in GUI understanding and navigation (fine tuned on smaller VQA Tasks). This model supports inputs at the resolutions of 1120x1120, enabling it to recognize tiny page elements and text in the flowcharts. 

\begin{table}[!htbp]
\centering
\small
\setlength{\tabcolsep}{1pt}
\begin{tabular}{lccc}
\toprule
\textbf{Open Model} & \textbf{LM} & \textbf{VM} & \textbf{Norm. Res.} \\
\midrule
CogAgent-VQA & Vicuna-7B & ViT-4.4B & 1120x1120 \\
{InternLM\textsubscript{-X-Comp.2}} & Intern-LM2-7B & ViT-304M & 490x490 \\
Qwen-VL-chat & Qwen-VL-7B & ViT-1.9B & 448x448 \\
\bottomrule
\end{tabular}
\caption{\small Open Baseline Models. VLMs are composed of a Language model that encodes text and a visual model that encodes the images. LM: Language Model, VM: denotes vision model.}
\label{table:base-models-stats}
\vspace{-1.0em}
\end{table}

\textbf{InternLM-X-Composer2}~\cite{DBLP:journals/corr/abs-2401-16420} uses a novel approach (PLORA) that applies additional LoRA parameters exclusively to image tokens to ensure that linguistic abilities are not affected, striking a balance between precise vision understanding and text composition. 
\textbf{Qwen-VL-chat}~\cite{Qwen-VL} is the instruction tuned model in the Qwen-VL series. Its \emph{position-aware vision language adapter} ensures that, even though the images are resized to a fixed resolution long image feature contexts are captured effectively by the model. We summarize the base language models and visual models used in our baselines in Table~\ref{table:base-models-stats}.

\subsection{Baseline Evaluation}
\label{subsec:VLM-evaluation}

We evaluate the baseline models under multiple settings:
\vspace{0.15em}
\begin{enumerate}[noitemsep, leftmargin=*, nolistsep]
    \item \textbf{Zero-Shot}:  Given a flowchart, we prompt the VLM to answer the question with a small instruction and provide a short concise answer.
    \item \textbf{Zero-Shot CoT}:  Given a flowchart, we prompt the VLM with the question to first elicit a rationale and then deduce the final answer \cite{wei2023chainofthought}.
    \vspace{0.15em}
    \item \textbf{Text Only Few-Shot CoT with Reasoning Directives}:  We create a custom prompt outlining the reasoning steps involved in answering questions specific to flowcharts. We scrutinize the areas where improved prompting is necessary for the models and draw inspiration from \cite{zhang2023multimodal}, \cite{li2023guiding}, and \cite{kojima2023large} to devise a text-only few-shot CoT approach with directional stimulus and step-by-step reasoning. The central objective is to deconstruct complex questions, identify which elements to map, and determine the answer. Each example, or "shot," encompasses four key components: The Question, Directional Stimulus Tags, Step-by-Step Rationale, and the Answer. These distinct parts aid in breaking down the question into relevant segments, offering a logical, step-by-step analysis, and concluding with an answer. We develop this strategy based on its potential effectiveness for flowcharts, with its actual efficacy demonstrated ahead. The few-shot samples we give are dynamic in nature, i.e the each question type gets more similar samples from our train set annotated samples samples for the method.
    \vspace{0.25em}
    \item \textbf{Fine-Tuning}:  We fine-tune the VLM on the train split of FlowVQA, and then prompt the VLM to answer the question.\footnote{Due to resource constraints we only Fine-Tune on Qwen-VL-Chat through LoRA Finetuning}
\end{enumerate}

\subsection{Evaluation Method}
Our methodology adopts an \textit{"AI as an Evaluator"} approach similar to \citet{fu2023gptscore, lin2023llmeval, chiang-lee-2023-large}. We employ three evaluator models—GPT-3.5 \cite{ye2023comprehensive}, Llama-2 70B \cite{touvron2023llama}, and Mixtral 8*7B (Mixtral-of-Experts) \cite{jiang2024mixtral} —to assess the model-generated responses, which are compared against three gold standard short answers and the question (context excluded). The evaluators' task is to dissect and align the responses, eliciting a detailed rationale that demonstrates Chain of Thought behavior, and then assigning a binary label to indicate whether the response is correct or incorrect. This process essentially boils down the evaluation into a "length-invariant" paraphrase detection task for short text responses, surpassing traditional similarity metrics and rule-based matching in effectiveness. We determine the final label via a majority vote among the evaluator models.
\paraendspace

\begin{table*}[!htbp]
\centering
\small
\setlength{\tabcolsep}{3pt} 
\renewcommand{\arraystretch}{1.15}

\begin{tabular*}{\textwidth}{@{\extracolsep{\fill}} llcccccccc}
\toprule
\textbf{Model} & \makecell{\textbf{Strategy}} & \makecell{\textbf{MV\textsubscript{Total}}} & \makecell{\textbf{MV\textsubscript{T1}}}  &  \makecell{\textbf{MV\textsubscript{T2}}} & \makecell{\textbf{MV\textsubscript{T3}}} & \makecell{\textbf{MV\textsubscript{T4}}} & \makecell{\textbf{MV\textsubscript{Wiki}}} &\makecell{\textbf{MV\textsubscript{Instruct}}} &\makecell{\textbf{MV\textsubscript{Code}}} \\

\midrule
\multirow{3}{*}{\shortstack{\textbf{GPT-4V}} } & \textbf{Zero-Shot }& 61.22 & \textbf{90.72}\textsuperscript{*} & 82.24 & 63.79 & 40.62 & 60.98 & 60.78 & 62.65\\
& \textbf{Zero-Shot COT} & 65.57 & 72.79 & 69.94 & 73.50 & \textbf{58.25}\textsuperscript{*} & \textbf{67.84}\textsuperscript{*} & 70.89 & 47.71\\
& \textbf{Few-Shot COT\textsubscript{D}} & \textbf{68.42}\textsuperscript{*} & 89.02 & \textbf{89.92}\textsuperscript{*} & \textbf{81.41}\textsuperscript{*} & 46.72 & 63.33 & \textbf{72.25}\textsuperscript{*} & \textbf{64.83}\textsuperscript{*} \\
\midrule
\multirow{3}{*}{\shortstack{\textbf{Gemini-Pro-V}} } & \textbf{Zero-Shot} & 49.57 & 80.08 & 70.29 & 35.34 & 33.86 & 48.84 & 48.27 & 54.36 \\
& \textbf{Zero-Shot COT} & 58.76 & 81.21 & 78.39 & 62.14 & 41.99 & 54.23 & 57.57 & 63.81 \\
& \textbf{Few-Shot COT\textsubscript{D}} & 61.41 & 84.96 & 81.83 & 77.69 & 43.60 & 54.12 & 60.12 & 61.41 \\
\midrule
\multirow{3}{*}{\shortstack{\textbf{CogAgent-VQA}} } & \textbf{Zero-Shot} & 37.17 & 55.27 & 52.68 & 26.56 & 27.23 & 37.45 & 36.80 & 36.96  \\
& \textbf{Zero-Shot COT} & 38.84 & 58.73 & 57.95 & 27.51 & 26.98 & 40.01 & 37.47 & 37.64 \\
& \textbf{Few-Shot COT\textsubscript{D}} & 25.13 & 33.93 & 34.26 & 16.76 & 21.67 & 34.62 & 29.65 & 22.37\\
\midrule
\multirow{3}{*}{\shortstack{\textbf{InternLM\textsubscript{-X-Comp.2}}} } & \textbf{Zero-Shot} & 37.47 & 49.47 & 49.79 & 24.16 & 32.15 & 35.67 & 38.26 & 41.90 \\
& \textbf{Zero-Shot COT} & 43.35 & 58.85 & \textbf{65.58}\textsuperscript{\#} & 33.86 & 31.39 & 43.24 & 41.48 & 47.16 \\
& \textbf{Few-Shot COT\textsubscript{D}} & 45.09 & 58.96 & 64.80 & 38.56 & 32.64 & 45.05 & \textbf{43.03}\textsuperscript{\#} & \textbf{47.74}\textsuperscript{\#}\\
\midrule
\multirow{3}{*}{\shortstack{\textbf{Qwen-VL-chat}} } & \textbf{Zero-Shot} & 33.67 & 48.83 & 46.64 & 20.19 & 26.89 & 32.92 & 34.02 & 35.47 \\
& \textbf{Zero-Shot COT} & 36.19 & 49.84 & 53.82 & 22.65 & 28.13 & 36.01 & 35.41 & 38.32 \\
& \textbf{Few-Shot COT\textsubscript{D}} & 38.44 & 57.21 & 57.00 & 25.13 & 27.98 & 40.76 & 37.75 & 32.94\\
\midrule
\textbf{Qwen-VL-chat} \textsubscript{FT} & \textbf{Zero-Shot} & 36.84 & 56.95 & 49.86 & 25.75 & 25.77 & 39.64 & 34.63 & 32.51\\
& \textbf{Zero-Shot COT} & \textbf{47.13}\textsuperscript{\#} & \textbf{61.55}\textsuperscript{\#} & 59.78 & \textbf{43.34}\textsuperscript{\#} & \textbf{36.02}\textsuperscript{\#} & \textbf{50.10}\textsuperscript{\#} & 42.14 & 47.67 \\
\bottomrule
\end{tabular*}
\caption{\small Majority Vote Accuracy on All Models and Strategies broken down Question Type Wise (\textit{T1, T2, T3, T4}) as in Sec~\ref{subsec:qa-creation} and Source-Wise (Instruct, Wiki, Code) as in Table~\ref{table:source-wise-stats}. The highest value for each column is highlighted and marked with * in Closed Source Models and with \# in Open Source Models.}
\vspace{-1.0em}
\label{tab:baseline-eval}
\end{table*}

\textbf{Fine-tuning Settings}. We fine-tune Qwen-VL-chat \textsubscript{FT} using LORA~\cite{DBLP:conf/iclr/HuSWALWWC22} strategy on 2xNVIDIA A100 40GB GPUs. We train with an effective batch size of 8 using a cosine-based learning scheduler with a warmup. We set a higher warmup to ensure no loss of pretraining knowledge in the base model.

\subsection{Baseline Results and Discussion}
Table~\ref{tab:baseline-eval} tabulates the results of model evaluations across multiple strategies, with the scores split across various question types and text sources. Figure~\ref{fig: Experimental-results} in Appendix~\ref{apex:Results} provides a horizontal bar chart that compiles the results from the table.
\paraendspace

\textbf{FlowVQA is sufficiently hard.} The dataset resource presents a challenging task, with all the models. The evaluations highlight a scope for improvement for all the models. Our Best performing model with the top performing strategy, i.e. GPT-4 prompted with Few-shot directive-based prompting achieves 68.42\% Majority voting across all the evaluators.
\paraendspace

\textbf{Few-Shot Directives are helpful}. In the evaluation of most of our models, we observe that text-only few-shot CoT with reasoning directives outperforms other prompting strategies.  We observe 7\% improvement in GPT-4 evaluation and 12\% improvement in Gemini-Pro with this strategy.
CogAgent-VQA , however does not show an improvement with few-shot directives. We observe in our initial experiments that it was unable to generate directives and hence it could not make use of reasoning directives.
\paraendspace

\textbf{Proprietary models perform better than open-source models}. We observe that proprietary models heavily outperform the open-source models. GPT-4 with few-shot directives outperforms Qwen-VL-chat by a significant 30\%. 
\paraendspace

\textbf{Fine-tuning helps}. We fine-tune Qwen-VL-chat and evaluate by prompting with Zero-Shot and Zero-Shot CoT strategies. We see an improvement of 3\% from Zero-Shot prompting and 11\% improvement from Zero-Shot COT. This improvement emphasises the lack of flowchart understanding in original pretraining mixtures of these VLMs. The improvement in T2, T3 and T4 (~10\%) being more significant than T1 (5\%), can be attributed to the fact that fact-retrieval is a simpler task and does not need in-depth understanding of the flowchart structure. The fine-tuned model outperforms all other existing open-source models, which highlights the fact that \textit{FlowVQA} can be effectively used to introduce visual logic and reasoning in existing VLMs. 
\paraendspace

\textbf{Question Types.} We present the question-wise metrics in Table~\ref{table:base-models-stats}. It is evident from the table that all models consistently perform better on \textit{Fact Retrieval (T1)} and \textit{Applied Scenario (T2)} based based questions than on \textit{Flow-Referential (T3)} and \textit{Topological (T4)}. Outlined in Sec.~\ref{subsec:qa-creation}, \textit{T3} and \textit{T4} question types require thorough understanding of the flowchart and complex reasoning over the visual modality. 
\begin{table}[!htbp]
\small
\centering
\begin{tabular}{cc}
\toprule
\textbf{Number of Nodes} & \textbf{Average Accuracy}\\
\midrule
0-8 & 51.73 \\
8-17 & 45.74 \\
17-26 & 44.60 \\
35-44 & 38.99 \\
26-35 & 40.35\\
\bottomrule
\end{tabular}
\caption{\small Number of Nodes comparison (Average across all models and strategies). Performance decreases as number of nodes increases.}
\label{table:breakdown-nodes}
\vspace{-1.0em}
\end{table}

\textbf{Number of Nodes}. Using the Mermaid.js scripts, we obtain the count of nodes in each flowchart. We categorize the flowchart by binning the number of nodes present in them. A Large number of nodes implies a more complex representation of visual information, and hence the flowchart is harder to reason upon. The results in the Table~\ref{table:breakdown-nodes} confirms this fact. Figure~\ref{fig:perf-num-nodes} in Appendix~\ref{apex:Results} shows the decline of performance of models with increase in number of nodes.

\subsection{Directional Bias}
To study \textbf{\textit{RQ4}}, we parse the mermaid scripts of the FlowVQA flowcharts and systematically invert them to produce a inverted flowchart \textbf{"Bottom Top"} set. Bottom Top analysis helps further evaluate the Visual and Sequential nature of our resource. The Bottom Top Flowcharts look directionally counter-intuitive with the start nodes at the bottom and end at the top. We perform this inversion on 1,500 flowchart-question pairs on which all evaluators evaluate to "True" (correct response for all). We evaluate a the top-performing models and strategies obtained in Section~\ref{subsec:VLM-evaluation} on the inverted flowchart set to detect any presence of directional bias in the VLMs. 
\paraendspace

Table~\ref{table:directional-bias} highlights the fact that our best performing models do \textit{suffer from a directional bias} in understanding and reasoning over flowcharts. We see a significant 15\% drop in majority voting accuracy thorough with GPT-4. 
\begin{table}[!htbp]
\centering

\small
\begin{tabular}{lcc}
\toprule
\textbf{Model (Strategy) } & \textbf{\textbf{Top-Down}} & \textbf{Bottom-Up} \\
\midrule
\textbf{GPT-4V} \textsubscript{(CoT)} & 100.00 & 85.71  \\
\textbf{Qwen-VL-chat} \textsubscript{(CoT)} & 100.00 & 76.09  \\
\bottomrule
\end{tabular}
\caption{\small Directional Bias test, we evaluate on two models using CoT approach on 1500 flowchart-QA pairs. }
\label{table:directional-bias}
\vspace{-1.0em}
\end{table}

\textbf{Analysis}. The directional bias evaluation underlines an important lacking of existing VLMs. They suffer from biases introduced in pretraining mixture and do not ground their inferences in the context images which leads to a significant drop in their evaluation performances. Strategies like augmenting pretraining mixtures with counterfactual examples might help alleviating these issues, which we leave for future study.

\section{Related Work}

Vision language models have made large progress in diverse vision-language applications~\cite{Qwen-VL, DBLP:journals/corr/abs-2310-03744, DBLP:journals/corr/abs-2402-12185} with multiple benchmarks being proposed to aid effective evaluation of visual and textual grounding capabilities of these models. The MMMU benchmark \cite{yue2023mmmu} is designed to assess the model's inherent "subject-specific" knowledge and reasoning abilities across various subjects (such as Technology, Humanities, Health, and more). 

Benchmarks like TextVQA and DocVQA \citep{singh2019vqa, mathew2021docvqa,8953217, 10.1007/978-3-030-58558-7_30, NEURIPS2022_11332b6b, hudson2019gqa} evaluate the models' fine-grained transcription abilities on low-resolution images. More complex multimodal reasoning tasks, such as MathVista \cite{lu2024mathvista}, examine the models' abilities to integrate visual and mathematical logic. Benchmarks focusing on spatial multimodal reasoning include ChartQA \cite{masry2022chartqa, DBLP:journals/corr/abs-2312-15915, DBLP:conf/wacv/MethaniGKK20} and InfographicVQA \cite{mathew2021infographicvqa}. ChartQA is aimed at evaluating straightforward chart understanding and analysis, while InfographicQA poses direct logical questions about data visualizations and charts.

{\bf Prior Flowchart Works}. To our knowledge, there exists a study on Flowchart QA \cite{flowchartqa}, that suffers from major limitations. (i) Synthetically generated flowcharts with randomized scripts, (ii) Primarily poses structural questions and (iii) Uses multiple choice-based questions to evaluate weaker existing models.
Other research in this domain addresses issues like Flowchart Object Recognition and Flowchart to Code/Script conversion, where a modest parallel flowchart resource is paired with corresponding code or script \cite{liu-etal-2022-code, 10.1007/978-3-031-41734-4_31, Thean2012TextualSO, 9795274}. However, notable limitations here include poor flowchart image quality, niche or overly complex context, structural imbalance (only linear or excessively complex), lack of ground truth scripts for flowcharts, and insufficient context for effective Q/A or practical tasks. In contrast to these works, we construct a complex benchmark suitable to test practical applicability of existing VLMs. The complex and diverse QA types ensure an effective and just evaluation over multi-modal visual and textual understanding.

\section{Conclusion and Future Work}
In conclusion, this study evaluates the effectiveness of existing Multimodal Large Language Models (VLMs) in reasoning upon a complex visual, sequential logical reasoning based task, \emph{FlowVQA}. We introduce the novel dataset resource, \emph{FlowVQA}, consisting of 2,272 Flowchart images, Mermaid.js scripts, 22,413 Q/A pairs with gold standard answers. Our extensive evaluation on these models with multiple strategies and scenarios highlights the need for advancements in \textbf{architecture} and \textbf{prompting} strategies in existing VLMs. 
We also study the presence of any \emph{directional} bias in the flowcharts by re-evaluating the test sets with an inverted flowchart subset. We find that both proprietary and open-source models suffer from directional bias due to lack of visual grounding and complex structural reasoning required for flowchart reasoning.

\paraendspace

\textbf{Future Work}. Our work and resources give rise to many research avenues in 
(a) \textbf{Flowchart Reasoning}: \emph{FlowVQA} can be used to enhance the visual logic and reasoning capabilities of the models. Constructing VLMs that are flowchart specific is also a encouraging research direction.
(b) \textbf{Graph-Encoder Models}: In this study, we consider the graph nature of flowcharts solely to generate topological questions. This consideration can also be taken into account while designing model architectures and inference strategies to enhance structural reasoning in the base models.
(c) \textbf{Adversarial and Counterfactual probes}: We provide questions of four different types which can be augmented with multiple probe sets like negative path following, counter-intuitive questions and noisy-graph based questions. 
(d) \textbf{Complex Subtasks}: The parallel nature of \emph{FlowVQA} allows us to formulate multiple subtasks using the resource. Primary task of \emph{FlowVQA} is the \textit{Flowchart$\rightarrow$Q/A}. We can create multitude of tasks:  \textit{article$\rightarrow$Q/A}, \textit{Mermaid.js$\rightarrow$Q/A}, \textit{Flowchart$\rightarrow$Mermaid.js}. The tasks can then act as an additional resource for training LLMs and VLMs.
(e) \textbf{NeuroSymbolic AI Approaches} like in \citet{Trinh2024} can also be considered to enhance performance and training on our resource as flowcharts are inherently symbolic and sequential structures.

\section*{Limitations}
There are a few notable limitations to our work. Primarily, the inability to fine-tune all models under consideration due to financial and computational resource constraints has led to a potential under-representation of the capabilities of various NLP models beyond our primary focus. Moreover, the language limitations encountered in this research, particularly the focus on English for generating Visual Question Answering (VQA) methods, underscore the need for linguistic diversity in NLP applications to ensure broader applicability and inclusivity. Given the novelty of the task at hand, it is also important to acknowledge that the insights provided may not be exhaustive, highlighting the potential for future research.

\section*{Ethics Statement}
We, the authors of this work, confirm that our research adheres to the highest ethical standards in both research and publication. Throughout this study, we have diligently considered and addressed various ethical issues to ensure the responsible and equitable use of computational linguistics methodologies. To promote the reproducibility of our results, we provide comprehensive information, including sharing code, datasets (we use publicly available datasets and comply with the ethical standards set by their authors), and other relevant resources. This enables the research community to validate and extend our work. The claims presented in this paper are consistent with our experimental results. However, given the inherent stochasticity of \textit{black-box} large language models, we have minimized variability by maintaining a fixed temperature. We thoroughly detail the annotations, dataset splits, models used, and prompting methods employed, ensuring the reproducibility of our work.

\section*{Acknowledgements}
Research was sponsored by the Army Research Office and was accomplished under Grant Number W911NF-20-1-0080. The views and conclusions contained in this document are those of the authors and should not be interpreted as representing the official policies, either expressed or implied, of the Army Research Office or the U.S. Government. The U.S. Government is authorized to reproduce and distribute reprints for Government purposes notwithstanding any copyright notation herein. This work was partially funded by ONR Contract N00014-19-1-2620. We extend our gratitude to the annotators who verified our flowcharts and corresponding question answer pairs. Lastly, we extend our appreciation to the reviewing team for their insightful comments.

\bibliography{custom, anthology, anthology2}
\bibliographystyle{acl_natbib}

\clearpage
\newpage

\appendix

\onecolumn

\section{Flowchart QA Example}
\label{apex:flowchart-qa-example}

\begin{tabular}{p{0.5\textwidth} p{0.5\textwidth}}
\raisebox{-0.4\totalheight}
{\includegraphics[width=0.5\textwidth]{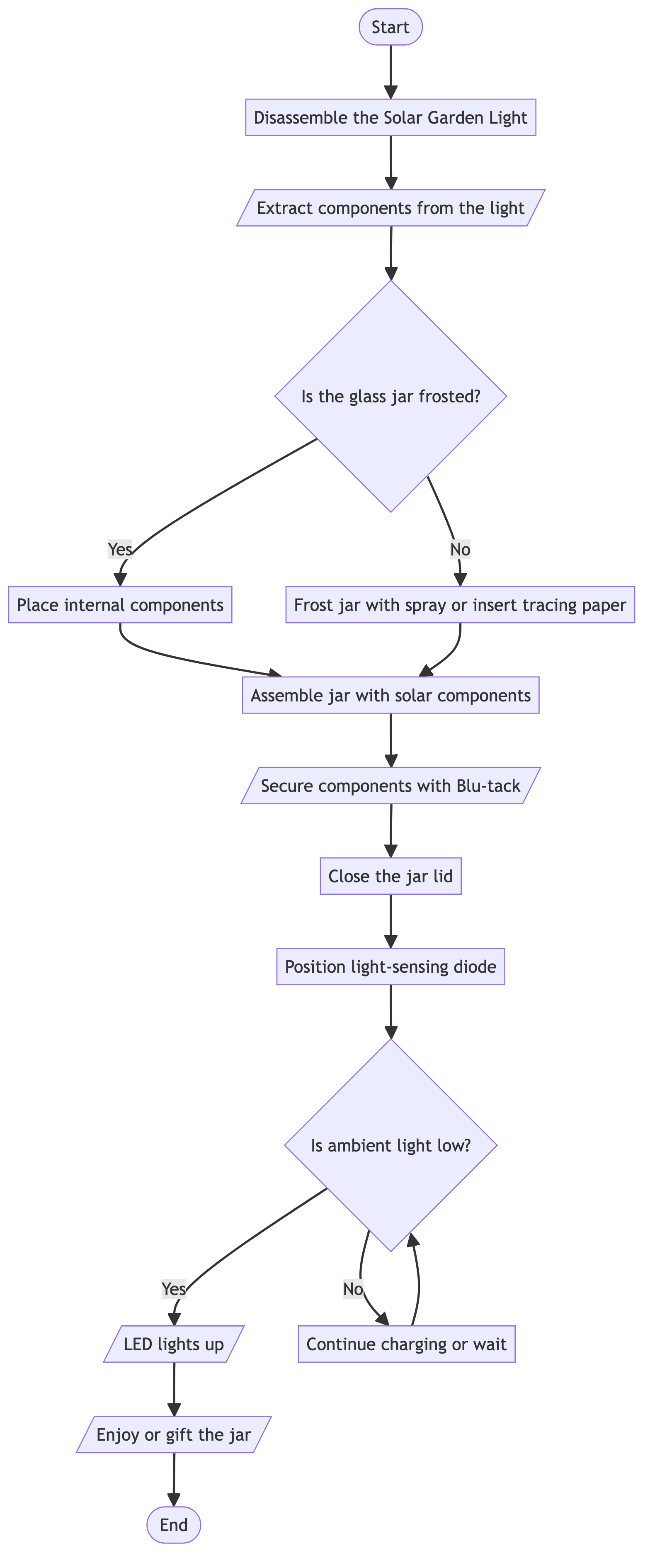}}
&
\begin{tabular}{p{\hsize}}
\textbf{T1: Fact Retrieval}\\
\textit{Q: What should be done if the glass jar is not frosted?} \\
\textit{A}: Frost the jar with spray or insert tracing paper. \\ \\ 
\textit{Q: What triggers the LED to light up in the solar jar?} \\
\textit{A}: Low ambient light causes the LED to light up. \\ \\ \\

\textbf{T2: Applied Scenario}\\
\textit{Q: Jason is disassembling a solar garden light for a DIY project but is unsure about how to safely extract the internal components including the solar panel, circuitry, LED, and battery housing. What tools should he use and how should he proceed with the disassembly?} \\
\textit{A}: Jason should use a utility knife and screwdriver to carefully disassemble the solar garden light and extract the necessary components. \\ \\ 
\textit{Q: While attempting to create a homemade solar-powered LED lighted cookie jar, Michael realized he forgot to frost his Ikea glass jar. He doesn't have any frosting spray on hand but remembers he has some tracing paper. How should he proceed to achieve the necessary frosted effect?} \\
\textit{A}: Michael should cut a strip of tracing paper to fit inside the jar to achieve the frosted effect.\\ \\ \\

\textbf{T3: Flow Referential}\\
\textit{Q: Assuming the glass jar was already frosted, what are the next two steps I must take in sequence?}\\
\textit{A}: You would place the internal components and then assemble the jar with solar components. \\ \\ 
\textit{Q: If I have just completed frosting the jar with spray or inserting tracing paper, what is the next immediate step in the process?}\\
\textit{A}: The next step is to assemble the jar with solar components. \\ \\ \\

\textbf{T4: Topological}\\
\textit{Q: How many nodes exist in the given flowchart?}\\
\textit{A}: 15 \\ \\
\textit{Q: Is the node "Is ambient light low?" direct predecessor of the node "Place internal components"?}\\
\textit{A}: No
\end{tabular}
\end{tabular}

\clearpage
\newpage

\section{Dataset Distribution}
\label{apex:dataset-distribution}

\begin{figure}[h]
    \centering
    \includegraphics[width=0.8\textwidth]{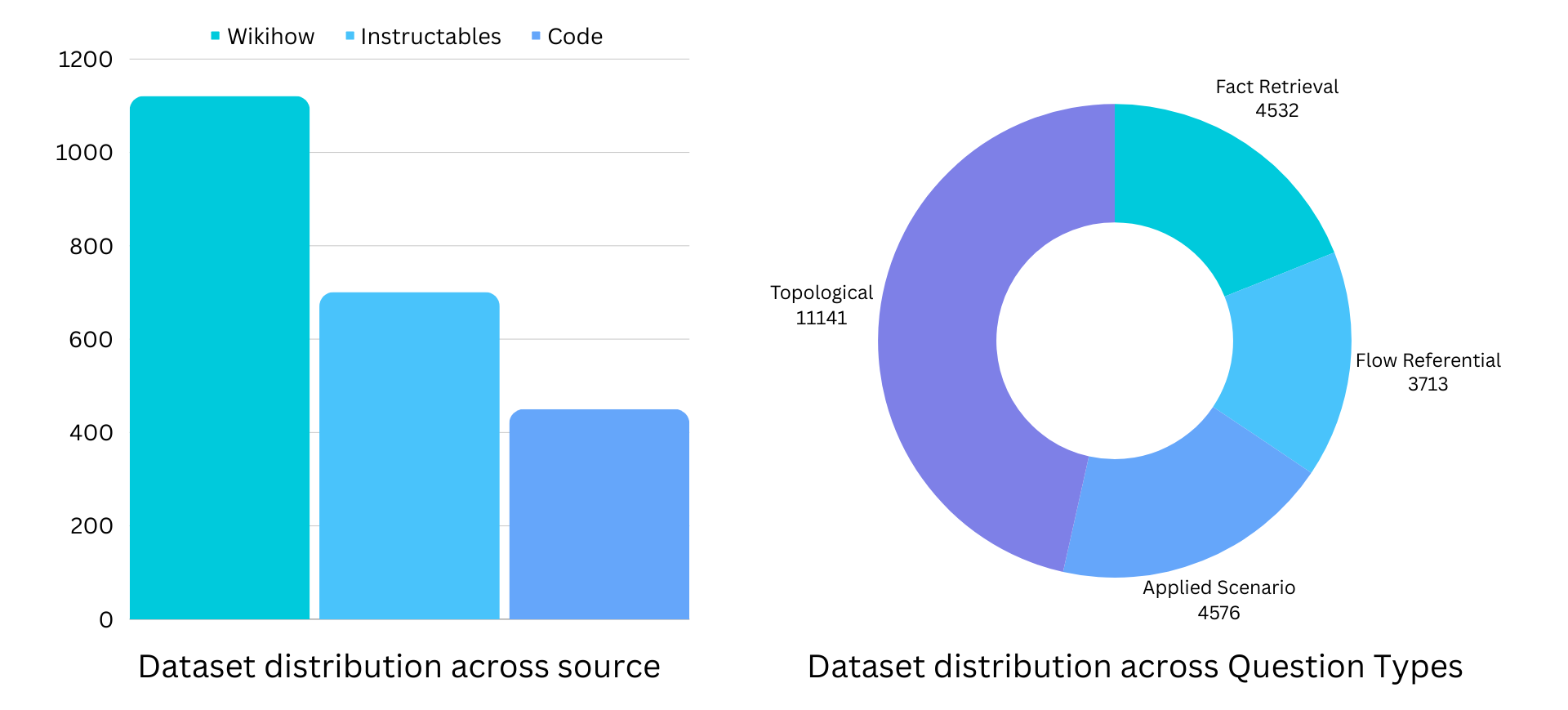}
    \caption{\small The figure shows the distribution of our data across different sources as well as across different types of questions.}
    \label{fig:dataset_distribution}
\end{figure}

Figure \ref{fig:dataset_distribution} illustrates how our data is distributed among various sources and types of questions.

\section{Additonal Results}
\label{apex:Results}
\begin{figure}[!h]
    \centering
    \includegraphics[width=\textwidth]{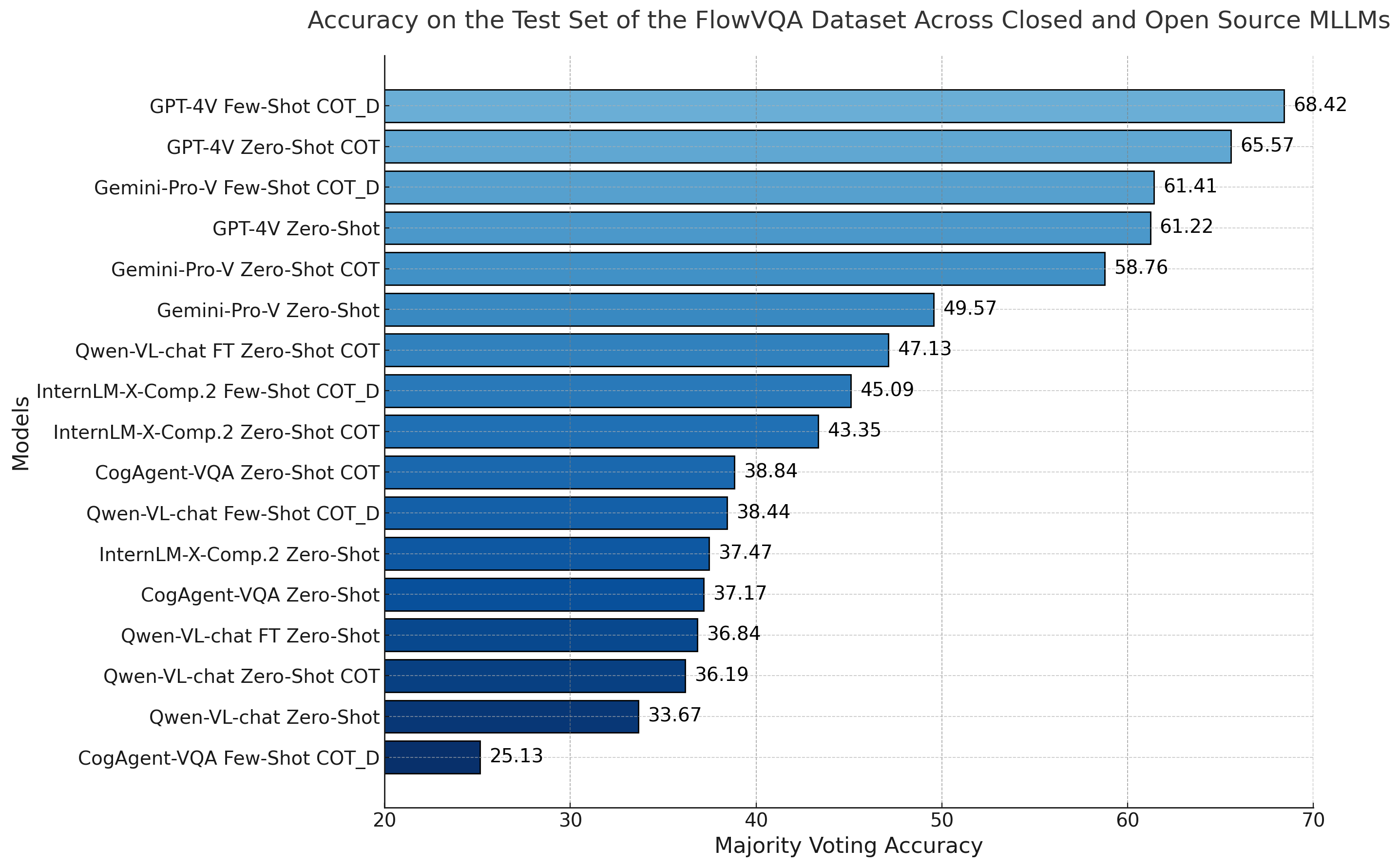}
    \caption{The horizontal bar chart shows the performance of FlowVQA dataset on various modelling strategies outlined in Section \ref{sec:experiment_evaluation}.}
    \label{fig: Experimental-results}
\end{figure}

\begin{figure}[!htbp]
    \centering
    \includegraphics[width=1\linewidth]{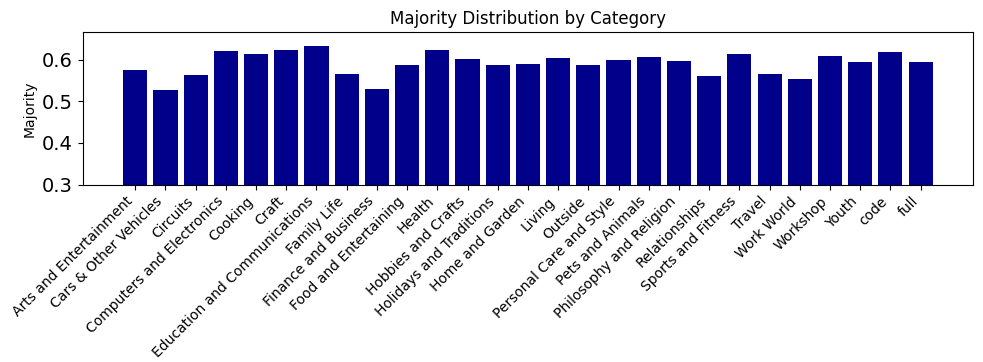}
    \caption{\small Category wise distribution of majority score for GPT-4V}
    \label{fig:Majority score}
\end{figure}

\begin{figure}[!htbp]
    \centering
    \includegraphics[width=\linewidth]{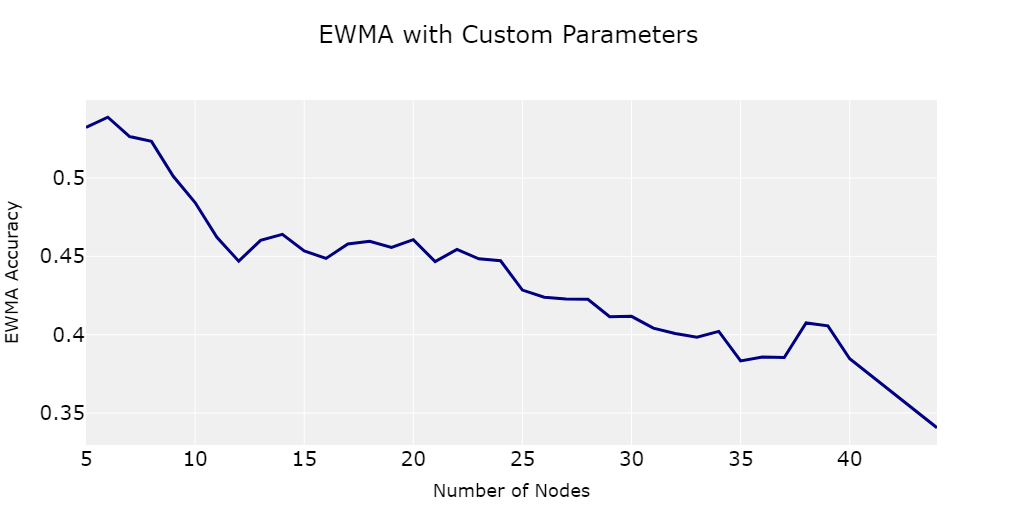}
    \caption{\small Average performance vs number of nodes. 
    We measure the average across all models and strategies and the grpah is created after smoothening with an exponential weighted moving average ($\alpha=0.4$)}
    \label{fig:perf-num-nodes}
\end{figure}

Figure \ref{fig: Experimental-results} 
shows the performance of FlowVQA dataset on various modelling strategies as outlined in Section \ref{sec:experiment_evaluation}. Table \ref{tab:category_wise}
shows VLMs across the three evaluator models - GPT, Llama and Mixtral over the various categories in the FlowVQA dataset. Figure \ref{fig:Majority score} show category wise distribution of majority score for GPT-4V model. We also measure the average performance vs number of nodes in the flowcharts . The average is across all models and strategies and the graph is created after smoothening with an exponential weighted moving average ($\alpha=0.4$), as shown in figure \ref{fig:perf-num-nodes}.

\begin{table}[!h]
\small
\centering
\begin{tabular}{c|c|c|c|c}
\hline
Category & Majority & GPT & LLAMA & Mixtral \\
& Voting & &  &  \\
\hline
Arts and Entertainment & 57.4 & 57.4 & 58.2 & 59.3 \\
Cars \& Other Vehicles & 52.8 & 54.3 & 53.6 & 53.4 \\
Circuits & 56.3 & 57.3 & 57.0 & 61.1 \\
Computers and Electronics & 62.1 & 61.8 & 61.4 & 63.6 \\
Cooking & 61.3 & 62.8 & 60.4 & 64.2 \\
Craft & 62.3 & 63.9 & 62.6 & 64.5 \\
Education and Communications & 63.4 & 64.4 & 60.2 & 64.4 \\
Family Life & 56.4 & 57.8 & 57.1 & 58.4 \\
Finance and Business & 53.1 & 54.6 & 53.4 & 53.8 \\
Food and Entertaining & 58.7 & 58.3 & 56.4 & 61.4 \\
Health & 62.4 & 64.8 & 60.4 & 62.8 \\
Hobbies and Crafts & 60.1 & 59.1 & 59.5 & 62.1 \\
Holidays and Traditions & 58.6 & 59.1 & 60.3 & 60.3 \\
Home and Garden & 59.0 & 59.8 & 57.0 & 60.5 \\
Living & 60.5 & 60.3 & 60.3 & 63.4 \\
Outside & 58.8 & 61.0 & 56.5 & 61.5 \\
Personal Care and Style & 59.9 & 59.9 & 60.3 & 62.9 \\
Pets and Animals & 60.7 & 62.1 & 60.3 & 64.7 \\
Philosophy and Religion & 59.6 & 58.2 & 58.7 & 60.9 \\
Relationships & 56.1 & 56.1 & 54.8 & 59.2 \\
Sports and Fitness & 61.3 & 62.9 & 60.2 & 60.9 \\
Travel & 56.6 & 57.0 & 55.3 & 58.3 \\
Work World & 55.3 & 54.3 & 53.2 & 56.7 \\
Workshop & 60.9 & 60.6 & 57.7 & 65.4 \\
Youth & 59.3 & 58.8 & 58.8 & 59.3 \\
code & 61.7 & 63.3 & 62.9 & 63.8 \\
\hline
\end{tabular}
\caption{\small Baselines across the three evaluator models—GPT, Llama and Mixtral over the various categories in the FlowVQA dataset.}
\label{tab:category_wise}
\end{table}

\newpage
\section{Prompts for Generation}
In this section we lists the prompts we use to query GPT-4 in various steps outlined in Section~\ref{sec:our-flowvqa-resource}

\subsection{Flowchart Creation}
\label{apex:flowprompts-for-generation}
\begin{table*}[!h]
\small
\centering
\begin{tabulary}{\textwidth}{LJ}
\toprule
\textbf{First Step: Breaking down source text to structured summaries} \\ 
\midrule

\textit{Please provide a comprehensive structured summary, detailed step-by-step representation of the blog post below. Each step in the representation summary should be labeled with specific control codes that define its nature in the system. These codes include:} \\ 

\textit{START}: Marks the first step. There must be only one start step and the whole summary representation must follow a single step-by-step structure. \\
\textit{PROCESS}: Indicates an ongoing process step. \\
\textit{DECISION [IF] [ELSE]}: Denotes a conditional decision-making step, with outcomes being either 'Yes' or 'No'. For steps with multiple outcomes, break them down into smaller decision steps. \\
\textit{INPUT}: Introduces new variables or elements, like ingredients in a recipe. \\
\textit{OUTPUT}: Highlights the results, outputs or products of a step \\
\textit{END}: Marks all terminal points where the process ends or cannot go any further. \\ 

! Treat the blog instructions as a system. The system has some inputs and some output. Describe the entire detailed summary in that particular format. Be it the working of an ATM machine or the steps to create pizza from raw ingredients everything can be looked at like a system or pseudocode. Make sure not to miss any critical points in processes. \\
! Try to retain context and structure it well. \\
! Important. Design the decision/conditional steps to have only 'Yes' or 'No' outcomes and treat their text like questions.\\
! Start from a single start point, do not have multiple parallel starts, make sure things remain step-wise with conditionals, loops etc.\\ 

\textit{Make the steps comprehensive and detailed, final output in markdown.} \\ 

\bottomrule
\end{tabulary}
\end{table*}

\begin{table*}[!h]
\small
\centering
\begin{tabulary}{\textwidth}{LJ}
\toprule
\textbf{Second step: Converting structured summaries to Mermaid Scripts} \\
\midrule

\textit{Here is a detailed step-by-step summary tagged with detailed control codes for a blog post. Treat the step-wise summary as a system or a detailed pipeline. For this create a \textit{Mermaid Live Flowchart Script (flowchart TD)} that is detailed, does not miss any key points, and captures all integral nodes perfectly. Treat the blog instructions and the flowchart as a system representation. Be it the working of an ATM machine or the steps to create pizza from raw ingredients everything can be looked at like a system.} \\ 

\textit{Objective}: Convert Passed Structured Summary to detailed Mermaid Live Flowchart (flowchart TD) \\ 

Control Codes for Assistance: \\
\textit{START}: Marks the first step. There must be only one start step and the whole summary representation must follow a single step-by-step structure. \\
\textit{PROCESS}: Indicates an ongoing procedure or action. Rectangle Shape. \\
\textit{DECISION [IF] [ELSE]}: Denotes a conditional decision-making step, with outcomes being either 'Yes' or 'No'. For multiple outcomes, decompose into smaller decisions. Diamond Shape. \\
\textit{INPUT}: Introduces new elements or variables, akin to ingredients in a recipe. Parallelogram Shape. \\
\textit{OUTPUT}: Results, Outputs or end-products of a step. Parallelogram Shape. \\
\textit{END}: All points of no further go terminal. Oval Shape. \\ 

Important Points \\
1. Treat the blog post instructions as a single system workflow or pipeline. \\
2. The system should include I/O, processes, decisions and terminals. \\
3. Ensure that the flowchart accurately depicts a real-life system flowchart, it should be contextually rich and practical for reference. \\
4. Maintain an optimal length for the flowchart not too long not too short, if there are multiple process steps in sequence you may consider combining them if the flowchart is too long. \\
5. Important! Design the decision steps to have only 'Yes' or 'No' outcomes. For steps with multiple outcomes, break them down into smaller decision steps. \\
6. Ensure a singular flow for the system, with all subroutines being direct components of the main system. \\
7. Ensure use of all flowchart symbols like rectangles, ovals, diamonds, circle, arrows etc. \\
8. Ensure the actual control codes are not mentioned in the flowchart nodes. \\
9. Verify flowchart syntax carefully \\ \\

\textit{Sample of a small mermaid flowchart TD for reference:} \\
flowchart TD \\
A(["Start"]) --> B["Process 1"] \\
B --> C{"Decision?"} \\
C -->|"Yes"| D["Process 2"] \\
D --> E["Process 3"] \\
E --> C \\
C -->|"No"| F[/"Output or Input"/] \\
F --> G(["End") \\ 

\textit{Make sure to verify each point above before your output.} \\ 
\bottomrule
\end{tabulary}
\end{table*}
\clearpage
\newpage

\subsection{Question Generation}
\label{apex:qgprompts-for-generation}
\begin{table*}[!h]
\centering
\small

\begin{tabulary}{\textwidth}{LJ}
\toprule
\textbf{Fact Retrieval} \\
\midrule
\textit{Task: You will analyze a step-by-step structured summary and \textit{Mermaid Flowchart Representation} of a blog post or code script. The blog post includes specific steps for handling tasks.} & \\ \\
\textit{Your Role}: As a fact-extractor and question creator, your objective is to locate factual content within the summary. Your goal is to construct several question-answer pairs that each relate to distinct and critical facts presented in the summary. & \\ \\

\textit{Guidelines for Question Development:} & \\
1. Begin by determining the presence and quantity of direct facts in the summary. If there are multiple concrete facts, especially quantitative ones, generate questions for each. If fewer facts are present, create fewer questions. The ideal question range is 2-4 questions. 2-3 for fewer facts and 3-4 for ones with more facts. & \\
2. Focus on specific and relevant facts, asking questions like Who? What? Why? How much? How many? Emphasize quantitative facts over qualitative ones.  & \\
3. Questions should be straightforward, with answers in the summary. Avoid direct references to the summary or the blog post in your questions. & \\
4. Ensure each question highlights a different fact from the summary. & \\ \\

\textit{Answer Guidelines:} & \\
1. Provide brief and clear answers. & \\
2. Answers must be definitive, avoiding open-endedness. & \\
3. Offer several paraphrased answers for each question. (A1, A2, A3) & \\ \\ 

\textit{Output Format:} Present your questions and answers in a structured JSON format, following the provided example. & \\
Example Structure: & \\
- Output JSON: & \\
\{ & \\
      "1": \{ & \\
        "Q": "First Fact-based Question here", & \\
        "A1": "", & \\
        "A2": "", & \\
				"A3": "", & \\
      \}, & \\
      "2": \{ & \\
        "Q": "", & \\
        "A1": "", & \\
        "A2": "", & \\
				"A3": "", & \\
			\}, & \\
\} & \\ \\

\textit{Sample Question-Answer Pairs:} & \\
1. What is the correct temperature for preheating the oven? \\ 
A1. 80 Degrees Celsius \\
A2. Preheat the oven to 80 degrees Celsius \\
A3. ... & \\ \\
2. How long should crayons be left in the oven to melt? \\ 
A1. 20 Minutes \\
A2. Leave the crayons in the oven for about 20 minutes & \\ \\
3. What might tempt someone to peek? \\
A1. Gifts \\
A2. The temptation to peek at Christmas gifts & \\ \\
4. At what angle should the target be struck for full extension? \\
A1. A 90-degree Angle & \\ \\
5. How long should the cork be left to cure? \\
A1. Overnight \\
A2. Cure the cork overnight & \\ \\

\textit{PS: Your Answers should be BRIEF, definitive and must offer three paraphrased versions A1, A2, A3. Make sure the questions are not too open ended and concrete.}  & \\ \\
Also DO NOT MENTION THE BLOG/STRUCTURED SUMMARY/SCRIPT IN THE QUESTION. & \\

\bottomrule
\end{tabulary}
\end{table*}

\clearpage
\newpage

\begin{table*}
\centering
\small
\begin{tabulary}{\textwidth}{LJ}
\toprule
\textbf{Applied Scenario} \\
\midrule
\textit{Task: You will analyze a step-by-step structured summary and Mermaid Flowchart Representation of a blog post or code script. The blog post includes specific steps for handling tasks.} \\ \\
\textit{Your Role}: As a complex situational question-answer generator, your task is to focus on the most interesting parts of the blog post's structured summary. Create 2-4 Complex Question-Answer Pairs. Each pair should correspond to a different, interesting area of the structured summary of the blog post. \\ \\

\textit{Guidelines for Question Development:} \\
- Focus on specific, relevant / crucial steps of the structured summary such as decisions, loops and other critical steps. \\
- Craft situational questions that are creative, practical, and likely to occur in real life. \\
- Ensure each question is directly related to a specific step mentioned in the blog post summary. \\
- Important: The question must be created in a way that the answer to the question can be directly obtained or inferred from the structured summary but no logical thinking should be done to further process the information in steps. The blog post should only be used to construct the context of the situation, not to generate the question itself. \\
- Important: Don't explicitly mention the structured summary or blog post in the question. Assume the person answering can reference it. Create long complex situations and questions. \\
- Provide suitable distractors in the question, complex stories, unique names, etc. Anything that makes the question more interesting, yet, answerable. \\
- Make sure all questions attend separate parts of the structured summary. \\ \\

\textit{Answer Guidelines:} \\
- Provide short, concise answers. \\
- Answers should be definitive and not open-ended. \\
- Offer several paraphrased answers for each question. (A1, A2, A3) \\ \\

\textit{Output Format:} Present your questions and answers in a structured JSON format, following the provided example. \\
Example Structure: \\
- Output JSON: \\
\{
      "1": \{ \\
        "Q": "First Applied Scenario Based Question", \\
        "A1": "Concise Answer 1", \\
        "A2": "", \\
				"A3": "", \\
      \}, \\
      "2": \{ \\
        "Q": "", \\
        "A1": "", \\
        "A2": "", \\
				"A3": "", \\
			\}, \\
				... More Q/A Pairs here \\
\} \\ \\

\textit{Sample Questions: }\\
1. Ram, aged 45 years old, was going home from the office in his Minivan and his Minivan broke down on the way. He now wants to find a Minivan mechanic to get it repaired. He was trying to follow the given article, but being a little forgetful, he could not remember the age of his Minivan. He thought his warranty documents could help, Where should he try to find them? \\
2. Alice has decided to make custom fabric paint for a set of cotton t-shirts. She mixed equal parts of acrylic paint and a transparent gloss medium, but after testing on a swatch of cotton, the paint soaked through. What adjustment should she make to the paint mixture? \\
3. Selena has recurrent tonsil stones and her doctor has prescribed a course of antibiotics to address the issue. Unfortunately, the antibiotics weren't successful and Selena hasn't experienced any side effects or a relapse. What would her doctor's advice likely be at this stage? \\
4. Mark, an aspiring VFX artist, is enthusiastic about networking to enhance his opportunities in the field. He wants to join an industry group like the Visual Effects Society (VES). However, he is uncertain about the number of VES members and their global distribution. How can Mark find this information to ensure the group's relevance to his networking goals? \\ \\

\textit{PS: Your Answers should be BRIEF, definitive and must offer three paraphrased versions A1, A2, A3. Make sure the questions are not too open ended and concrete.}  \\ \\
Also DO NOT MENTION THE BLOG/STRUCTURED SUMMARY/SCRIPT IN THE QUESTION. \\
  
\bottomrule
\end{tabulary}
\end{table*}
\clearpage
\newpage

\begin{table*}
\centering
\small
\begin{tabulary}{\textwidth}{LJ}
\toprule
\textbf{Flow Referential} \\
\midrule
\textit{Task}: You will analyze a step-by-step structured summary and Mermaid Flowchart Representation of a blog post or code script. This post details specific steps to handle certain tasks. \\ \\
\textit{Your Role}: As a capable flowchart path and flow analyzer your task is to focus on critical sub-areas of the processes and flowchart and create path-based questions from that subflowchart.  \\ \\

\textit{Guidelines for Question Development:} \\
- The first step is to decide on how many questions to create: If the flowchart is long and complex, break it down into smaller areas and create more questions (3). If the flowchart is short create fewer (2-3) but still good quality questions that would not be easy to answer directly. Focus on specific, relevant / crucial paths of the structured flowchart script and summary.  \\
- Create questions based on node information looking FORWARD, BACKWARD, IN THE MIDDLE, etc. Questions about crucial decisions taken in a possible path.  \\
- Craft questions about paths that are creative and hard but MUST HAVE A SINGLE DEFINITIVE TRUE ANSWER.  \\
- Important: Don't explicitly mention the structured summary or flowchart in the question. Assume the person answering can reference it. Create long complex situations and questions.  \\
- Create questions about backtracking, future paths, conditionals, nodes or steps in the middle, etc. Anything that is interesting in a flowchart path.  \\
- IMPORTANT! It is very important that the current node/step or the node/path in question later is mentioned clearly. The rules for counting must be clearly mentioned.  \\
Look at the sample questions below to create questions.  \\ \\

\textit{Answer Guidelines:} \\
- Provide concise direct answers that are relevant to the question asked.  \\
- Answers should be definitive.  \\
- Offer several paraphrased answers for each question. (A1, A2, A3)  \\ \\

\textit{Output Format:} Present your questions and answers in a structured JSON format, following the provided example.  \\
Example Structure:  \\
- Output JSON: \\
\{ \\
  "1": \{ \\
    "Q": "First Path Based Question", \\
    "A1": "Concise Answer 1", \\
    "A2": "", \\
    "A3": "" \\
  \}, \\
  "2": \{ \\
    "Q": "", \\
    "A1": "", \\
    "A2": "", \\
    "A3": "" \\
  \}, \\
\} \\

\textit{Sample Questions:}  \\
1. What is the second step, given my zeroeth step is taking a negative decision at "Bostik Spritzkork 3070 Available?"?  \\
2. If I currently have to fill the mold with plaster, what decision must have I taken a few steps back and what is the condition present at that node?  \\
3. What is the minimum number of steps required to reach 'Final Inspection' from the "change job?" conditional?  \\
4. Given the current zeroeth step is to close the top of the lid, what is the fifth step that I will be completing if I take the affirmative decision at any conditional present in between?  \\
5. If at the current step the bathtub is not yet full and requires more water, what are the labels or descriptions of the fifth and seventh steps encountered when following the affirmative path from the current decision node?  \\
6. How many steps are there from the initial "Start" node up to, but not including, the first decision point? In this count, the "Start" node is to be considered as the initial node or the 'zeroeth' step.  \\
7. Alice is preparing for a rock-themed party and recalls Scarlet's unique style. She decides to start with a band T-shirt but is unsure whether to buy it online or at a concert. Given her limited budget, what should Alice's decision be based on?  \\
8. If a patient's eligibility for tonsillectomy is currently being evaluated and they proceed with tonsillectomy following a positive recommendation, what would be the immediate next step, and what decision must have been made directly prior to this step?  \\ \\

\textit{Answers:}  \\
9. If I am currently at the 'Choose Show Audio Animation or press Control-A' step, what was the decision made at the first decision point, and what is the immediate next step?  \\
    A1: "The decision made was 'Yes' at the 'Decision to edit audio effects?' node, and the immediate next step is 'Audio effects editing mode activated'.  \\
    A2: "At the 'Decision to edit audio effects?' node, a positive decision was taken, leading to the next step of activating the audio effects editing mode.  \\
    A3: "The first decision point led to a 'Yes' outcome, and the following step is to activate the audio effects editing mode.  \\ \\

\textit{PS: Your Answers should be BRIEF, definitive and must offer three paraphrased versions A1, A2, A3. Make sure the questions are not too open ended and concrete.}  \\
Also DO NOT MENTION THE BLOG/STRUCTURED SUMMARY/ FLOWCHART SCRIPT IN THE QUESTION.  \\
  
\bottomrule
\end{tabulary}
\end{table*}

\clearpage
\newpage

\section{Prompts for Question-Answering}
In this section we lists the prompts we use to query models
\subsection{Few-Shot-COT-with-Directives}
\begin{table*}[h]
\centering
\small
\begin{tabulary}{\textwidth}{L}
\toprule
\textbf{Few-Shot COT\textsubscript{D}} \\
\midrule
Examine the provided flowchart to answer the given question below. Here are some illustrative examples accompanied by a sequence of reasoning directives intended to stimulate analytical thought and elicit a rationale. These guidelines should facilitate the development of a rationale. Once a rationale has been formulated, proceed to present a conclusive final answer as in the examples. \\

\textit{Question - Answer pairs with Tags. The exemplar questions given depend upon the type of question we are asking} 
\\\textbf{(Fact Retrieval/Applied Scenario/Flow Referential/Topological)}
\\Example:\\
$\vdots$ \\
Q1. What temperature should the oven be preheated to for making the cake?
\\
Tags: Temperature, Oven, Preheating, Cake
Reasoning: Take it step-by-step. Look for a node or cluster of nodes, in the flowchart that mention preheating the oven. Identify node / nodes that mention a 'preheating'.
After locating relevant nodes extract final answer if already present or reason further to deduce the correct answer.
\\
A. 325 degrees Fahrenheit.\\
$\vdots$ \\\\
\textit{Concerned Question to Ask} \\
Example:\\
What action is taken when the 'file' is found to be not empty? \\
\bottomrule
\end{tabulary}
\end{table*}

\clearpage
\newpage

\end{document}